\documentclass[11pt,a4paper]{llncs}
\usepackage{amsmath}
\usepackage{amssymb}
\usepackage{graphicx}
\usepackage{marvosym}
\usepackage{url}
\usepackage{fancyhdr}
\usepackage{subcaption}

\usepackage{geometry}
\newcommand{\keywords}[1]{\par\addvspace\baselineskip
\noindent\keywordname\enspace\ignorespaces#1}

\fancypagestyle{firstpage}{\fancyhf{}
}

\begin{document}

%\mainmatter  % start of an individual contribution

\title{\LARGE{A transition towards virtual representations of visual scenes}}

%\titlerunning{Lecture Notes in Computer Science: Authors' Instructions}

% \author{\large{Afg Cer  \and A. Bcd \and P. Qwe \and R. Bcgfee \and R.Y. Tedf}}
% \institute{\large{School of XYZ and RST, University of ABCDEF,\\ Qwergh-12345, XYZ}}

\author{\large{Américo Pereira\inst{1,2} \and Pedro Carvalho\inst{1,3} \and Luís Côrte-Real\inst{1,2}}}
%\cortext[cor1]{Américo Pereira: Tel. xxxxxxxxxxx;}
% \emailauthor{americo.j.pereira@inesctec.pt}{Américo Pereira}
% \author[1,3]{Pedro Carvalho}
% \author[1,2]{Luís C\^{o}rte-Real}
\institute{\large{Centre for Telecommunications and Multimedia, INESC TEC, Porto, Portugal \and 
Faculty of Engineering, University of Porto, Porto, Portugal \and 
Polytechnic of Porto, School of Engineering, Porto, Portugal}}

%\author{Alfred Hofmann%
%\thanks{Please note that the LNCS Editorial assumes that all authors have used
%the western naming convention, with given names preceding surnames. This determines
%the structure of the names in the running heads and the author index.}%
%\and Ursula Barth\and Ingrid Haas\and Frank Holzwarth\and\\
%Anna Kramer\and Leonie Kunz\and Christine Rei\ss\and\\
%Nicole Sator\and Erika Siebert-Cole\and Peter Stra\ss er}

%\institute{Springer-Verlag, Computer Science Editorial,\\
%Tiergartenstr. 17, 69121 Heidelberg, Germany\\
%\url{http://www.springer.com/lncs}}


%\toctitle{Lecture Notes in Computer Science}
%\tocauthor{Authors' Instructions}

\maketitle

\thispagestyle{firstpage}

\begin{abstract}
Visual scene understanding is a fundamental task in computer vision that aims to extract meaningful information from visual data. It traditionally involves disjoint and specialized algorithms for different tasks that are tailored for specific application scenarios. This can be cumbersome when designing complex systems that include processing of visual and semantic data extracted from visual scenes, which is even more noticeable nowadays with the influx of applications for virtual or augmented reality. When designing a system that employs automatic visual scene understanding to enable a precise and semantically coherent description of the underlying scene, which can be used to fuel a visualization component with 3D virtual synthesis, the lack of flexibility and unified frameworks become more prominent. To alleviate this issue and its inherent problems, we propose an architecture that addresses the challenges of visual scene understanding and description towards a 3D virtual synthesis that enables an adaptable, unified and coherent solution. Furthermore, we expose how our proposition can be of use into multiple application areas. Additionally, we also present a proof of concept system that employs our architecture to further prove its usability in practice.
\end{abstract}

%\begin{abstract}
%The abstract should summarize the contents of the paper and should
%contain at least 70 and at most 150 words. It should be written using the
%\emph{abstract} environment.
%\keywords{We would like to encourage you to list your keywords within
%the abstract section}
%\end{abstract}

%%%%%%%%%%%%%%%%%%%%%%%%%%%%%%%%%%%%%%%%%%%%%%%%%%%%%%%%%%%%%%%%%%%%%%%%%%%%%%%%%%%%%%%%%%%%%%%%%%%%%%%%%%%%%%%%%%%%%%%%%%%%%%%%%%%%%%%%%%%%%%%%
\section{Introduction}
Visual scene understanding is a fundamental task in computer vision that aims to extract rich and meaningful information from visual data. It plays a crucial role in numerous real-world applications where perception and interpretation of visual information is required in order to assess and complete different tasks. 
Nowadays, with the increased attention towards virtual reality, augmented reality and overall interest in providing richer forms to visualize data, it becomes clear that there is a need to integrate 3D techniques and methods with visual scene understanding. Hence, the task of automatic visual scene understanding for 3D scene synthesis can be seen as a new challenge. This involves automatic perception, analysis and interpretation of visual data that can be employed into a dynamic 3D scene through the usage of multiple sensors and algorithms. This new challenge can see application in multiple application scenarios, such as: surveillance, sports, retail or entertainment. As an example, in~\cite{cui2023fusing} visual data and synthesis are used to create a mixed reality system that allows users to explore a 3D environment.  

\begin{figure*}[t!]
    \centering
    \includegraphics[width=0.75\linewidth]{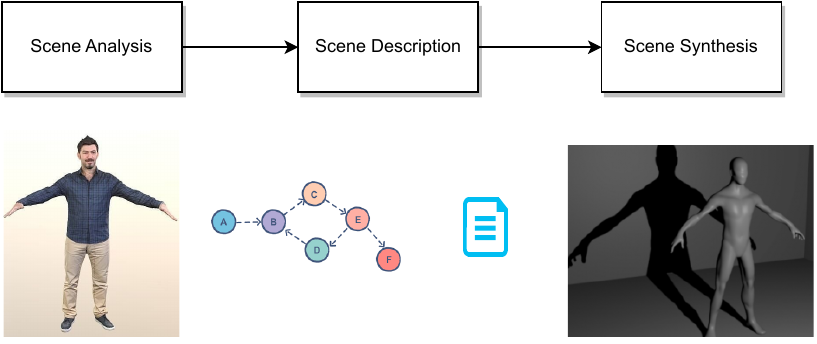}
    \caption{Visual-virtual translation pipeline proposed in~\cite{pereira2023visualscene}.}
    \label{fig:initial_architecture}
\end{figure*}

Traditional approaches to scene understanding often involve separate and specialized algorithms for different tasks, leading to fragmented and disjointed analysis that hinders the system's ability to achieve a holistic and coherent understanding of visual scenes. As observed in~\cite{pereira2023visualscene}, there is evidence of a need for a well structured and unified framework that is capable of analyzing a scene, describing and synthesizing it. This could provide several advantages over traditional disjointed approaches, such as allowing for a seamless integration of different modules, facilitating information exchange and enabling synergy among tasks. Visual scenes are composed of diverse objects, spatial relationships, contextual cues, and temporal dynamics, which collectively contribute to the overall understanding; thus, it becomes necessary to formulate a cohesive framework that enables a comprehensive and contextually-aware understanding of the visual scene by leveraging different types of information. In such a framework it is important that the knowledge extracted from the visual scene is accessible through the entire processing chain, as multiple algorithms that are part of the framework may use this information to enhance the overall understanding of the scene. Contextual understanding is another crucial aspect that a unified framework can address, as scenes are not merely a collection of objects but are characterized by spatial layout, temporal dynamics and semantic coherence. 

In this article we propose an architecture that enables the creation of a unified framework or system that addresses the challenges of visual scene understanding for 3D scene synthesis. To achieve this, we start by leveraging the initial basic architecture proposed in~\cite{pereira2023visualscene}, depicted in Fig.~\ref{fig:initial_architecture}, and expand the modules, detailing aspects of the framework. Our proposal consists of four main components: scene analysis, scene description, scene synthesis and a data orchestrator. In the scene analysis module, visual input is processed to extract low and high-level features, detect objects, infer semantic segmentation, estimate poses, and capture contextual relationships among objects. This information serves as the foundation for subsequent stages, facilitating a detailed understanding of the scene. The scene description module takes the output of the scene analysis and constructs a high-level representation of the scene that incorporates the spatial information, object attributes, semantic labels and contextual information to generate a structured scene description. Finally, the scene synthesis module utilizes the scene description to generate a realistic and immersive 3D representation of the scene. This generation can blend semantic data, spatial arrangement, and time-based elements to create realistic scenes for specific purposes, offering adaptable and flexible solutions based on input restrictions and output needs. The data orchestrator is responsible to ensure a common ground on all of the processes and concepts within the system, effectively guaranteeing consistency of the data flow across the entire architecture. It also includes an important sub-module that helps in the creation of an informed decisions on the best algorithm combinations to be applied to the input data.

The contributions of this work are three-fold: 1) we showcase the new challenge of visual scene understanding for 3D scene synthesis; 2) we present an unified and flexible system architecture to take on this challenge; and 3) we show a practical application of this architecture by implementing a proof of concept system that incorporates our designs and provide examples of generated hybrid scenes that can be obtained by our system, illustrating the capability of generating synthetic data that could be used to train other models. We also present a series of possible applications that could leverage our proposal to target specific problems.

The document is structured as follows: Section~\ref{sec:related} explores existing works in the field of visual scene understanding and discusses existing methodologies, algorithms, and frameworks used in scene analysis and synthesis. Section~\ref{sec:architecture} presents the proposed unified architecture in detail, providing an overview of the architecture as a whole and explaining the role of each component and their interactions within the system, exploring possible technologies and algorithms that can be applied in each component. Section~\ref{sec:usecases} delves into potential use cases and areas of application that benefit from employing our proposal. Section~\ref{sec:experiment} presents a proof of concept system that incorporates the main ideas of our proposal into a system and exemplifies possible outcomes that can be obtained. Finally, section~\ref{sec:conclusions} summarizes the contributions of the article and discusses future research opportunities and directions required to further improve the proposed system architecture. 

%%%%%%%%%%%%%%%%%%%%%%%%%%%%%%%%%%%%%%%%%%%%%%%%%%%%%%%%%%%%%%%%%%%%%%%%%%%%%%%%%%%%%%%%%%%%%%%%%%%%%%%%%%%%%%%%%%%%%%%%%%%%%%%%%%%%%%%%%%%%%%%%
\section{Related Work}\label{sec:related}
With the improvement of processing power and neural network design, several areas of scene understanding have naturally evolved, with the proposal of new methods and more detailed datasets. When looking at image recognition, works such as NFNets~\cite{brock2021high} show that it is possible to achieve high accuracy on large image datasets such as ImageNet~\cite{deng2009imagenet} with a faster training process. RepMLP~\cite{ding2021repmlp} shows that incorporating prior information into fully connected layers enhances image recognition abilities. Video object segmentation has also evolved, with works such as:  SwiftNet~\cite{wang2021swiftnet} that uses pixel-adaptive memory and pixel-wise memory update and match to reduce temporal and spatial redundancy, enabling real time processing; and LCM~\cite{hu2021learning}, which also uses a memory-based approach into a semi-supervised method that addresses the problem of not using the sequential order of the frames and object-level knowledge. Another related topic in visual scene understanding is salient object detection; the work presented in~\cite{wang2021salient} studied and compared several approaches, ultimately concluding that there are still many under-explored problems in achieving efficient and reliable network designs. In a recent work, the algorithm ID-YOLO~\cite{qin2022idyolo} is proposed to achieve real-time salient object detection by extending the well known YOLOv3~\cite{redmon2018yolov3} algorithm with the instance segmentation algorithm Poly-YOLO~\cite{hurtik2022polyyolo}.   

\begin{figure}
    \centering
    \includegraphics[width=0.5\linewidth]{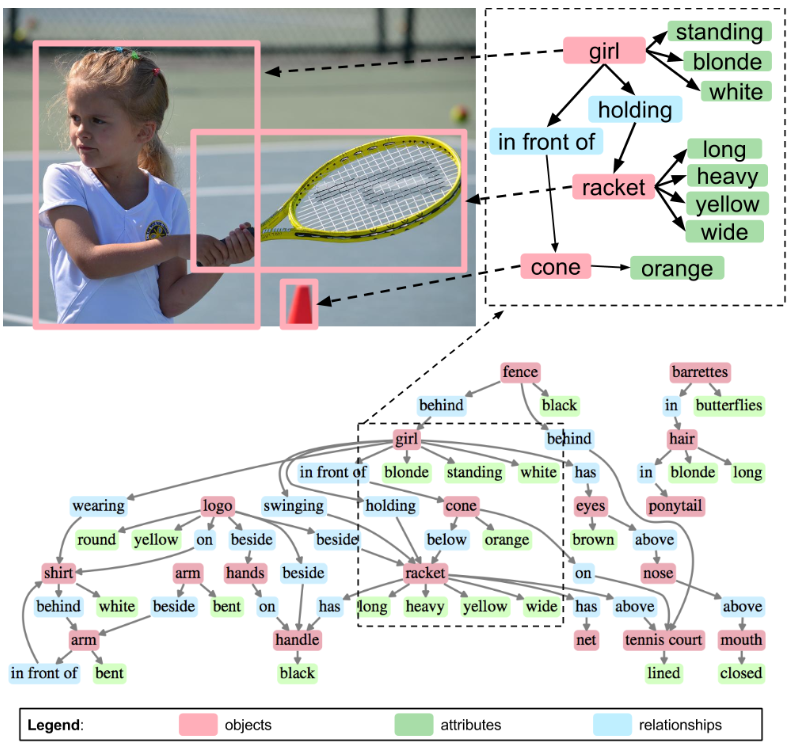}
    \caption{An example of a scene graph (taken from~\cite{johnson2018image}).}
    \label{fig:sg_example_1}
\end{figure}

Considering an hierarchical perspective over a scene, detection and tracking can be seen as starting points of a more complete understanding of the information present. Hence, more high-level subjective aspects, such as the meaning of the location of the objects, activities or even the interactions that occur are important and, therefore, a semantic parsing of visual scenes is necessary. A way to amass and convey these details extracted from a visual scene is through the usage of a Scene Graph; which is a data structure that is mainly used to describe objects, attributes and their relationships. It can represent the semantic details of a scene by explicitly modeling objects along with their attributes and relationships. They were originally introduced in Johnson et al.~\cite{johnson2015image} and, since then, research on their generation and application to multiple scenarios has progressed. Scene graphs have been used for tasks such as image/video captioning~\cite{gao2018image,yang2019autoencoding,kim2019dense}, visual question answering~\cite{koner2021graphhopper,nuthalapati2021lightweight,cherian2022spatio}, image retrieval~\cite{nguyen2021deep,schroeder2020structured} or image generation~\cite{mody2022analysis,xue2022deep}. Despite the research interest in scene graphs, most of the existing works are related to generating the graphs from single images. In the case of videos, there are approaches that use spatio-temporal scene graphs to model the semantic information present in the sequence; however, due to the  constraints that are introduced due to temporal observations, the process of generating the scene graph becomes increasingly difficult. Works such as~\cite{tsai2019video,malawade2022spatio} try to use state-of-the art video object detection and tracking methods to generate the graphs and the results obtained are starting to become more accurate. Fig.~\ref{fig:sg_example_1} depicts an example of a scene graph, where objects, attributes and relationships represent the semantic information of the image. 

Human activity detection is also a very challenging and studied topic where the improvements of computational capabilities and neural networks enabled considerable advances. In~\cite{bilal2022transfer}, an RNN with LSTM was used to learn long-term temporal relationships in order to achieve spatio-temporal human action recognition in long videos that have overlapping actions. In a different field, the work presented in~\cite{brehar2021pedestrian} detects street-crossing pedestrians for a safer autonomous driving system. Multiple state-of-the-art works are also explored in detail in~\cite{patil2022survey}, where action recognition algorithms are compared in multiple application scenarios. Pose estimation is also a powerful tool to assess human activity and in~\cite{song2021human} 2D skeleton-based action recognition methods that estimate the pose of humans from RGB images are compared and assessed. Analogous is a study for 3D skeleton-based action recognition~\cite{ren2020survey}.

When looking at works that specifically mention visual scene understanding, it is noticeable that it is viewed mostly as a fixed concept. For instance, in~\cite{zhou2023mtanet} RGB and thermal images are used on a multitask-aware network that mixes semantic information with coarse features at various abstraction levels to ultimately segment images. In~\cite{hu2020probabilistic} a deep learning framework for future video prediction is presented, where the authors incorporate a module for scene understanding that serves to reconstruct semantic segmentation and depth images and predict optical flows. A bidirectional projection network is proposed in~\cite{hu2021bidirectional} to leverage the complementary information of 2D and 3D data to provide, once again semantic segmentation, but for 2D and 3D. Semantic scene completion is another related topic, where a 3D scene is reconstructed by leveraging visual and semantic data extracted from single-view depth or RGBD images~\cite{Cai_2021_CVPR,Cao_2022_CVPR}. There is also work on end-to-end semantic instance reconstruction from incomplete point clouds~\cite{Nie_2021_CVPR,behley2021towards}. These works ultimately show that visual scene understanding has a vast area of application. However, there is a tendency to link the concept towards semantic segmentation and not to the more general idea of extracting semantic data from visual scenes.  

As with other areas, 3D virtualization has also evolved in recent years. In particular, human parametric models have been used for multiple scenarios such as 3D human pose and shape estimation~\cite{wang2022human}, controllable 3D human synthesis~\cite{xu2022surface} or virtual try-ons of clothing~\cite{vidaurre2020fully}. There has also been research on using graph convolutional neural networks to generate 3D human shapes with better resolution~\cite{yu2022multi}. Also, 3D modeling tools and game engines such as Unity~\cite{unity3d}, Unreal Engine~\cite{unrealengine4} or Blender~\cite{blender} have also evolved, introducing new features and more support for new graphic interchange languages such as Universal Scene Description (USD)~\cite{baillet2018forging}, which is a universal format for 3D graphics. However, its usage for 3D reconstruction algorithmic pipelines is still extremely uncommon, showing that its inclusion will lead to new research opportunities.  Nowadays, with the usage of Generative Adversarial Networks (GANs)~\cite{ferreira2022gan,fu2021single,shi2022deep} or Neural Radience Fields (NeRF)~\cite{gao2022nerf,rabby2023beyondpixels}, there has been a significant increase in the realism of the generated images and 3D representations. However, there is still a lack of usability when integrating a less restrictive and versatile application scenario. 

%%%%%%%%%%%%%%%%%%%%%%%%%%%%%%%%%%%%%%%%%%%%%%%%%%%%%%%%%%%%%%%%%%%%%%%%%%%%%%%%%%%%%%%%%%%%%%%%%%%%%%%%%%%%%%%%%%%%%%%%%%%%%%%%%%%%%%%%%%%%%%%%
\section{Unified Architecture}\label{sec:architecture}
\begin{figure*}[t!]
    \centering
    \includegraphics[width=0.7\linewidth]{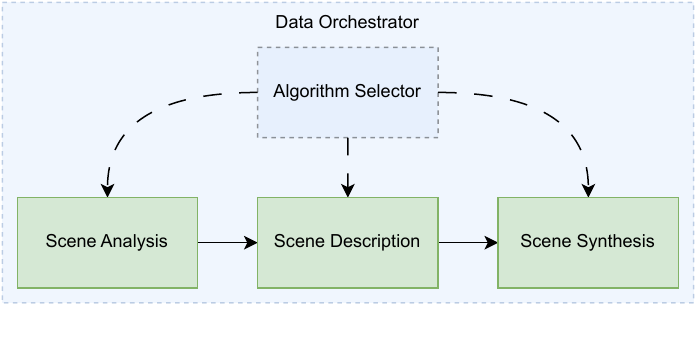}
    \caption{Proposed high level architecture for transitioning from visual scene towards 3D synthesis. We can see the three main modules depicted as green boxes and the two supporting components that glues the architecture as blue boxes.}
    \label{fig:architecture}
\end{figure*}

Embracing a broader vision of visual scene understanding than what is generally found on literature, we target the paradigm of visual scene understanding for 3D scene synthesis and explore the idea of a unified architecture that targets the processes required to transit from visual to semantic data, and further to a posterior 3D reconstruction. 
The proposed architecture, depicted in Fig.~\ref{fig:architecture}, employs four essential modules: scene analysis, scene description, scene synthesis and a data orchestrator. One support module is also present. In essence, the scene analysis module processes visual data, extracting key information like object detection and spatial relationships. The scene description module then constructs a high-level representation of the scene, capturing attributes and contextual details. Finally, the scene synthesis module uses this information enabling the creation of flexible, customizable and realistic 3D scene, incorporating semantic data, spatial layout, and temporal dynamics for an immersive experience. The data orchestrator serves as a central hub for defining and sharing data, such as object types, attributes, and relationships. This way, the standardization of the knowledge domain promotes consistency and interoperability across the system, enabling seamless communication between components. The algorithm selector sub-module is a dynamic component that assesses user input and considers factors like scene type, complexity, resources, and desired output to then intelligently choose which scene analysis algorithms to run and what type of information needs to be include in the scene description.

By incorporating these components, we allow a system that implements this architecture to be able to dynamically adapt to different scenes and requirements, thus optimizing the performance and output of the system. In the following subsections we delve into each of these components and detail their internal structure and what type of data flow and algorithms could be used for their implementation.

\subsection{Algorithm Selector}
The Algorithm Selector intends to provide flexibility to the entire system. 
It receives user input information and is responsible to interpret this information and provide a selection of the most appropriate algorithms to be used to process the input video or images. The decision must take into consideration the type of scene that is to be analyzed, the type of information that is to be described and the desired output in order to effectively select the most appropriate group of algorithms. In the current proposal, the algorithm selector follows a rule-based selection process, but can evolve into a more sophisticated process. By making use of rule sets provided in the configuration, the Algorithm Selector is able to provide a deterministic selection of algorithms and associated parameter to process input visual data. This allows for fine-grained control over the algorithms so that they can target specific problems that can appear and are accounted for by the rule sets. 

\subsection{Data Orchestrator}
Visual scene understanding requires the integration of multiple algorithms, addressing different types of information that can be extracted from a scene. This means that for a system to incorporate an ensemble of algorithms, it needs to have a common ground where the type of data and the flow of information is well structured, thus enabling the adaptation of multiple algorithms and enabling future development without requiring structural re-definitions of the entire architecture. 
This can be critical, for instance, to reduce latency in critical applications such as self-driving cars or adjusting UAVs depending on their surroundings and sensory information. To address these issues, we propose the Data Orchestrator module, which binds and connects all components in the system. For this module we propose the usage of a knowledge base or ontology that provides a shared and consistent representation of objects, attributes, and relationships, ensuring a common understanding across algorithms. To achieve this, a structured representation of the knowledge base/ontology needs to be designed, so that it captures the semantics and relationships between different concepts in the scene understanding domain. 

The structured representation can be implemented using various technologies and standards such as RDF (Resource Description Framework)~\cite{McBride2004}, OWL (Web Ontology Language)~\cite{antoniou2009web}, or JSON-LD (JSON for Linked Data)~\cite{sporny2020json}. This can then be populated by integrating existing domain-specific knowledge sources, such as existing datasets like MS COCO~\cite{lin2014microsoft}, Open Image~\cite{krasin2017openimages} or Image Net~\cite{deng2009imagenet}; or external ontologies~\cite{olszewska2011ontology,kenfack2020robotvqa,amodeo2022og}. Algorithms within remaining modules can then access and query the knowledge base/ontology to obtain relevant information for their respective tasks.

\subsection{Visual Scene Analysis}
The Visual Scene Analysis module is the first step in the processing chain of the system. Visual input data in given to the module in the form of images or videos and the algorithms identified by the Algorithm Selector sub-module are applied to the data to extract meaningful information about the scene. As different types of information can be extracted, it employs a multi-stage processing pipeline to perform various tasks and ensure that processes required by multiple algorithms are not performed more than once.

\begin{figure}[t!]
    \centering
    \includegraphics[width=0.25\linewidth]{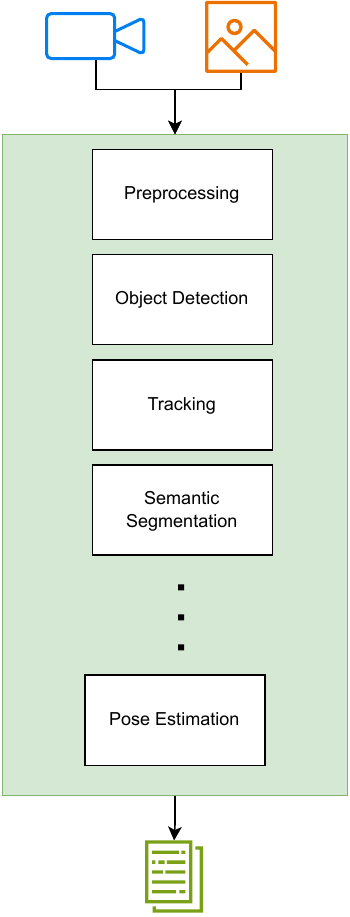}
    \caption{High level scheme of the visual scene analysis module. At the top we can have preprocessing techniques that are followed by more advanced processes to extract meaningful information. These processes may or may not be interconnected and depend on the application scenario and the rules specified as well as the desired output. The output of the module is processed data that is the forwarded to the scene description module.}
    \label{fig:scene_analysis}
\end{figure}

Internally, the module should have a hierarchical structure that goes from basic pre-processing techniques, such as noise reduction, image enhancements, normalization, resizes and other low-level image manipulation techniques; to more advanced techniques. Naturally, the processes required depend on the input data and on the requirements specified by the user and associated rule sets. These advanced techniques include processes such as: object detection and recognition, semantic segmentation, pose estimation, and so on. Additionally, other more specialized and advanced algorithms may also be integrated, based on the rules specified for the scene analysis algorithms. These may include scene classification, object tracking, action recognition, group behavior and so on. We can see in Fig.~\ref{fig:scene_analysis} an illustration of this module, where the hierarchy of the processing pipeline within the module can be observed. 

Overall, the internal architecture of the Visual Scene Analysis module incorporates a combination of traditional computer vision techniques and state-of-the-art deep learning-based methods to extract rich, high level and meaningful information from the input data. The combination of the results obtained by these algorithms is then forwarded to the Scene Description module to finalize the overall scene understanding process. 

\subsection{Scene Description}
The Scene Description module has two main responsibilities: enhance the analysis with semantic information in order to finalize the scene understanding process; provide means to describe the entire scene in a well structured way. With this in mind, the internal structure of the module is divided into two levels. One for finalizing the scene interpretation and another to prepare the obtained data for distribution. 
To achieve these responsibilities, different methodologies can be considered when we look at the specific requirements and constraints of a given application. For instance, relational databases could be used to store and manage scene data, which may be useful for applications that only require efficient retrieval and querying of data. Semantic networks~\cite{sowa1992semantic}, that represent objects or concepts as nodes, and semantic relationships between as edges are also useful as they can capture complex semantic relationships and dependencies within a scene. The ORA-SS (Object-Relationship-Attribute Data Model for Semi-structured Data) data model~\cite{Dobbie2009}, which focuses on representing objects, their attributes, and the relationships between them in a more tabular or relational format, can also be applicable for scenarios where the scene structure is less hierarchical. However, these methodologies have limitations if we consider arbitrary visual scenes and their inherent hierarchical structure that combine both visual and semantic data can be of use for other applications. Taking these limitations into consideration and the fact that we have a data orchestrator that provides a global ontology/knowledge base, we argue that incorporating scene graphs could be advantageous. The choice of using scene graphs comes from the fact that they are an hierarchical structure that captures the objects in the scene as nodes and their relationships as edges. Each node represents an object, and edges denote relationships between objects. Another important aspect is that scene graphs provide a compact and expressive representation that allows for rich scene understanding and reasoning. 

The module starts by extracting object-level and contextual information from the output of the Scene Analysis module, which includes: detected objects, their attributes (such as color, size, and shape), their spatial relationships, their interconnections within the scene, and then formulates a scene graph with these components. To enhance the accuracy and completeness of the scene description, the module leverages the structured knowledge within the ontology/knowledge base to correct potential errors or inconsistencies that may arise from the object detection or relationship detection processes by using a semantic reasoner to infer logical consequences from the data. It is important to highlight that there exist automatic scene graphs generation methods, which target both the scene analysis and description at the same time. Our idea is to incorporate these algorithms in the system, as well as enhance the description with other data extracted from the scene analysis. Works such as~\cite{cong2021spatial,xu2022meta,li2022dynamic}, are good candidates for this approach, as they leverage spatial and temporal information to better formulate and generate accurate scene graphs. Another good example is presented in~\cite{amodeo2022og}, where ontologies are used to refine the resulting scene graphs to better target specific application scenarios.

\begin{figure}
    \centering
    \includegraphics[width=0.5\linewidth]{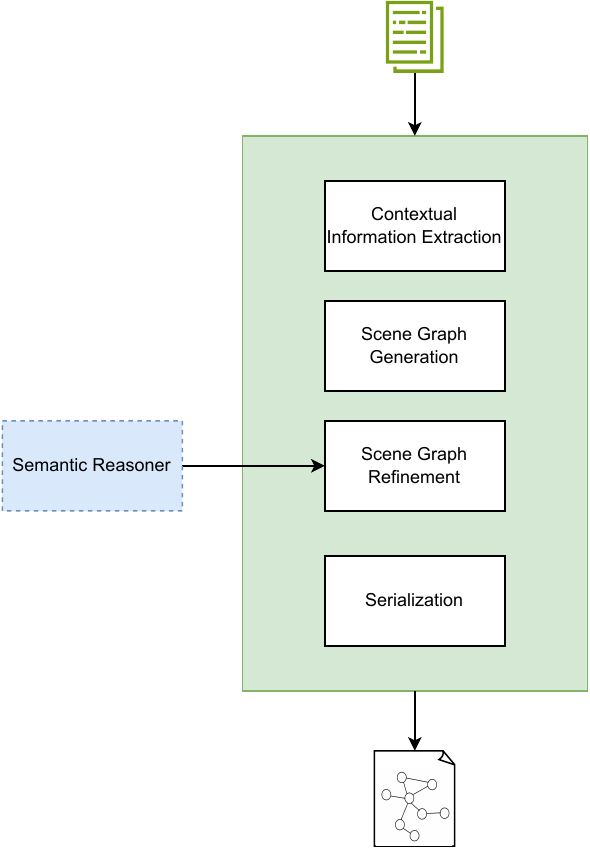}
    \caption{High level view of the scene description module. It receives the processed data from the scene analysis and extracts semantic information from it to then produce a scene graph. The output of the module is a serialized and refined graph.}
    \label{fig:scene_description}
\end{figure}

After the generation of the scene graph and its refinement, the module prepares a data structure to output the graph in a readable and editable way. This involves encoding the generated scene graph and associated visual and semantic scene information into a suitable format that can be edited and interchanged, thus ensuring that the information obtained from the analysis of the scene can be stored and transmitted. A depiction of this module can be seen in Fig.~\ref{fig:scene_description}. To create this textual representation, the module may utilize various data formats such as JSON, XML or other formats that are tailored to the specific needs of the system. For instance, it may also support a binary format that allows further information compression and faster read speeds. It is important that the chosen format ensures that the data is stored efficiently and that the scene information extracted and processed is preserved both in integrity and completeness, so that this information can be further used without the need to analyze the scene once again. 

Overall, this module has a pivotal role in this architecture, as it is crucial for providing a structured and semantically rich output of the complete analysis of a visual scene. The representation provided by the module also serves as a fundamental basis for further processing, visualization and storage or exchange of scene information. For instance, users can manually change the description to modify the underlying scene and therefore enable the synthesis of different scenes without changing anything in the processing pipeline. This capacity also enables the support of a wide range of application and tasks, by leveraging the rich information contained in the representation.  

\subsection{Scene Synthesis}

The Scene Synthesis module is the final component of our architecture and it takes as input the structured description of the scene provided by the previous module. It leverages this description in order to generate a visually and semantically coherent 3D representation of the described scene. To achieve this, the module should internally combine 3D modeling techniques, rendering algorithms, and synthesis methods to generate a realistic and comprehensive visual output.

To generate the virtual scene, the module follows a series of steps, that depend on both the scene description and the user input that specifies the desired output. In a first phase, it starts by instantiating 3D models of the objects and their respective attributes within the scene. To do so, it can either utilize 3D modeling techniques such as geometric modeling or voxel-based representations or even a template-based approach where a 3D basic 3D model of each component is fetched from a database of previously generated models. Then, these models are positioned and arranged on the virtual scene according to the spatial information that is presented on the description. Once the positioning and instantiation of the models is made, the module begins the rendering process, where the lighting, shading, and texture mapping is made to improve the quality of the virtual environment. Naturally, this process is also very dependent of the user input that is given, as different techniques can be applied depending on the desired outcome. For instance, ray tracing, simple rasterization, global illumination simulation, and many others are techniques that can be employed in this process in order to provide the desired realism. This shows that the architecture not only supports different types of components for the scene analysis but also for the scene synthesis. Thus, enabling different implementations depending on the application scenario. To further improve realism techniques such as generative adversarial networks (GANs) or Neural Radience Fields (NeRF) could also be applied.  

\begin{figure}
    \centering
    \includegraphics[width=0.25\linewidth]{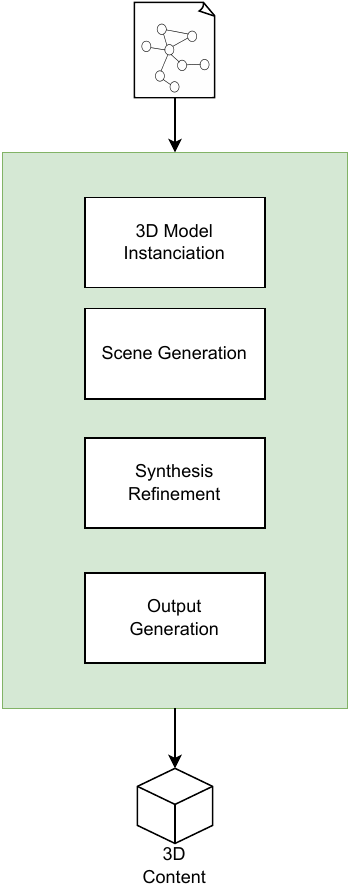}
    \caption{High level view of the scene synthesis module. It receives a serialized and refined graph and proceeds to translate the information present on the scene graph to 3D content that is then exported to a desired format.}
    \label{fig:scene_synthesis}
\end{figure}

The final output of the Scene Synthesis module is expected to be a visually appealing and coherent 3D representation of the analyzed scene, that ensures that the semantic information presented is the same as the one obtained from the scene analysis. This virtual representation should closely resemble the real-world environment described in the scene representation and follow the user instructions that define what type of output is to be created. This synthesized also has an important feature, where the representation can be rendered from different viewpoints and can have varying levels of detail, enabling visualization, virtual reality experiences, or integration into other applications. Fig.~\ref{fig:scene_synthesis} depicts a high level view of the scene synthesis module. 

The type and format of the output will also depend on specified user/application requirements, and can be of the following forms: rendered images, rendered videos, 3D models, USD, or other application dependent data formats. It could also export 3D models of the synthesized scene in various formats, such as FBX (Filmbox), OBJ (Wavefront Object), or COLLADA (COLLAborative Design Activity), that may include the geometry, textures, materials, and animations of the scene, allowing for further manipulation or integration with other software. Similarly, the usage of USD could also be enabled as it is an open and scalable interchange format for 3D scenes that enables efficient storage, sharing, and collaboration across different software tools and platforms. Finally, this architecture could also enable the integration of other different data formats that could be specifically tailored to other applications for a seamless integration, or even allow the generation of mixed reality scenes, where virtual components are included into a real scene.

%%%%%%%%%%%%%%%%%%%%%%%%%%%%%%%%%%%%%%%%%%%%%%%%%%%%%%%%%%%%%%%%%%%%%%%%%%%%%%%%%%%%%%%%%%%%%%%%%%%%%%%%%%%%%%%%%%%%%%%%%%%%%%%%%%%%%%%%%%%%%%%
\section{Possible Use Cases}\label{sec:usecases}

As we have seen above, our proposal puts forward the building blocks necessary to create a system or framework that enables visual scene understanding description and possible posterior 3D scene representation. As the workflow transitions from scene analysis to representation and then to synthesis, it can be applied in multiple areas that are related to each of these steps. In this section we will present some potential applications and use cases for a system that implements our architecture, discussing what can be achieved. 

As the last step of our proposal is the synthesis of virtual scenes, it is natural that one of the most direct use cases is related to content creation. 
By taking advantage of the possibility to pick a base scene and changed its description according to a user's desires, it is possible to synthesize 3D content for various purposes, such as movies, advertisements, or virtual worlds. Enabling content creators to generate realistic objects and scenes and without relying solely on physical setups or real-world recordings and manual labor creating the scenes from scratch. Furthermore, by having the possibility of controlling the synthesis process, it is possible to create different types of visualization of the same input data by enabling more or less detail or even providing different points of view. Similarly, it sees application in the creation and augmentation of datasets either of purely virtual spaces and actors, or mixed content, with virtual avatars in real scenes. 

Another relevant area that sees usability is virtual and augmented reality, where users could recreate physical scenes and turn them into immersive and interactable virtual worlds, that could be easily changed by editing their associated descriptions. Additionally, by porting these descriptions into an augmented reality device, we could create virtual overlays with information of objects onto the real world, thus providing a seamless and immersing AR experience. In a similar fashion, it can also be applied for gamification or serious games, where specific situations can be recreated and used to provide valuable input for patient treatment and rehabilitation. 

Another important area that sees advantages in using our proposal is in surveillance and overall security environments, where a complete visual scene understanding system for synthesis can be a great tool to provide multiple types of information. For instance, it could: detect anomalies by analyzing the detections of objects, people and their interactions; provide a synthesized representation of the space, so that events can be analyzed without infringing personal data laws, as only virtual representation of arbitrary human models is stored and provided; help better define the position of camera solutions to ensure better surveillance; provide a powerful tool to store and visualize past recordings without the need to store the original video.

By employing this architecture, we gain the ability to analyze, describe, and synthesize visual scenes with a high level of accuracy and realism that is derived from the applied algorithms. This enables us to understand the content of images and videos and represent scenes in a structured manner, and also providing flexible output options, allowing users to select the most suitable format for their needs. This way, it sees applicability in a wide range of applications encompassing multiple scenarios.

%%%%%%%%%%%%%%%%%%%%%%%%%%%%%%%%%%%%%%%%%%%%%%%%%%%%%%%%%%%%%%%%%%%%%%%%%%%%%%%%%%%%%%%%%%%%%%%%%%%%%%%%%%%%%%%%%%%%%%%%%%%%%%%%%%%%%%%%%%%%%%%
\section{Proof of concept}\label{sec:experiment}
In this section, we present a practical application of the proposed architecture for visual scene understanding and 3D synthesis, applied to a real-world scenario. In this implementation we target the specific problem of generation of hybrid data, that integrates real live scenes with virtual avatars performing actions based on real actors. To do so we detect and swap existing humans in videos with 3D avatars performing the same activities, ensuring the anonymity of the original actors, which is an important aspect when data protection specifications are highly restrictive. This system integrates our proposed architecture and makes use of state of the art technologies. In this implementation we show how each component of the system can be plugged, highlighting the seamless integration of the scene analysis, description, and synthesis modules. As this proof of concept intends to show the possibilities of the proposed architecture, we do not elaborate a complex ontology or define multiple rules. Rather, we focus on detecting, tracking and translating humans from videos to a virtual representation that is then synthesized for an output video, while also allowing manual change of the description in order to manipulate the outcome of the synthesis without interfering with the input data.  

In the scene analysis module we detect, segment and track humans in the input videos. This is achieved by relying on the state-of-the-art YOLO-v8, presented by Ultralytics~\cite{jocher2023yolo}. This way, we identify bounding boxes, segmentation masks and obtain IDs for each human in the scene. This information is then provided to the VIBE~\cite{kocabas2020vibe} pose estimation algorithm, so that the pose of each human in the scene is detected and mapped to SMPL models, which are versatile and realistic representation of human body shape and pose~\cite{loper2015smpl}. For the scene description, we formulate a simple scene graph that contains the corresponding ID of the associated person, as well as pose and SMPL data associated with them, for each frame. In this implementation we also store separately the corresponding segmentation masks that can be used to aid the inpainting process required to remove the original detections from the images. This information is then serialized to a JSON file, as we want to illustrate the ability to store and distribute information extracted from the analysis of the scenes. This also shows the ability of the system to divide the processing pipeline, where Scene Analysis and Description can first be performed, and the Scene Synthesis module can later use the generated description to generate the synthesis. For the rules established in this implementation, we simply select an inpainting process if the user wishes to swap the humans in the input video with the 3D avatars. Otherwise the avatars are overlapped in the resulting video. For the inpainting we used the E$^{2}$FGVI algorithm~\cite{liCvpr22vInpainting} as it shows impressive results. To illustrate the results obtained by our proof of concept with and without inpainting, we selected videos from the HMDB51 dataset~\cite{kuehne2011hmdb51}.

\begin{figure}
    \centering
    \begin{subfigure}{0.325\linewidth}
        \includegraphics[width=\linewidth]{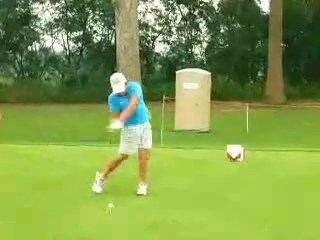}
    \end{subfigure}
    \begin{subfigure}{0.325\linewidth}
        \includegraphics[width=\linewidth]{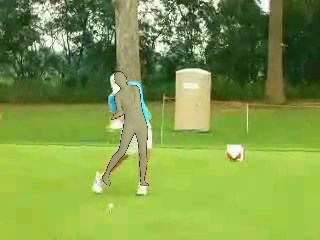}
    \end{subfigure}
    \begin{subfigure}{0.325\linewidth}
        \includegraphics[width=\linewidth]{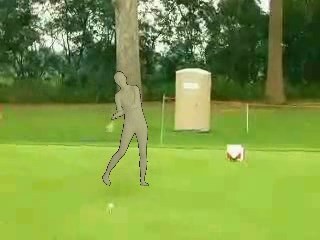}
    \end{subfigure}
    \caption{Example of a video of a woman swinging a golf club. In the center we have an avatar overlapping the original actor and on the right we completely removed the actor, resulting in a hybrid synthesis.}
    \label{fig:process}
\end{figure}

In fig.~\ref{fig:process} we show a simple example of the possible outcomes that can be obtained by our implementation. We can place avatars over the original video, or remove the original actor and place an avatar on its place. 

\begin{figure}[t!]
    \centering
    \begin{subfigure}{0.49\linewidth}
        \includegraphics[width=\linewidth]{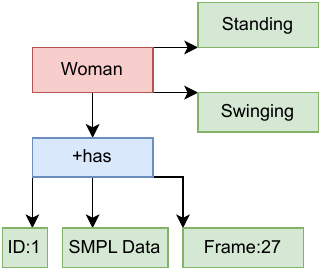}
    \end{subfigure}
    \begin{subfigure}{0.49\linewidth}
        \includegraphics[width=\linewidth]{images/golf_avatar_27.jpg}
    \end{subfigure}
    \hfill
    \begin{subfigure}{0.49\linewidth}
        \includegraphics[width=\linewidth]{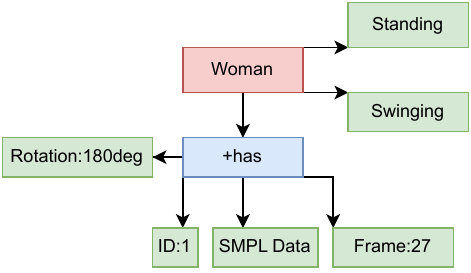}
    \end{subfigure}
    \begin{subfigure}{0.49\linewidth}
        \includegraphics[width=\linewidth]{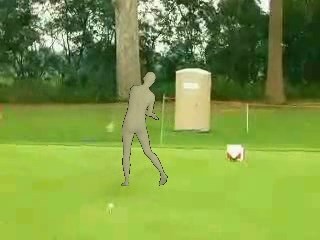}
    \end{subfigure}
    \caption{In the top images we have an avatar swinging a golf club, with the corresponding scene graph associated. On the bottom we see the same synthesis but with a modification in the scene graph, resulting in a rotation of the avatar on the synthesis process. We use red to denote entities, green for attributes and blue for relationships. }
    \label{fig:example_2}
\end{figure}

In the next example, depicted in Fig.~\ref{fig:example_2}, we show that for an arbitrary frame we can obtain a corresponding scene graph that represents the information in it. Furthermore, by manipulating the graph and introducing a rotation attribute, we can tell the Scene Synthesis module to apply a rotation on the generation of the avatar. In the graph, we use the special \textit{+has} relationship to include the data extracted from the analysis. 

\begin{figure}
    \centering
    \begin{subfigure}{0.325\linewidth}
        \includegraphics[width=\linewidth]{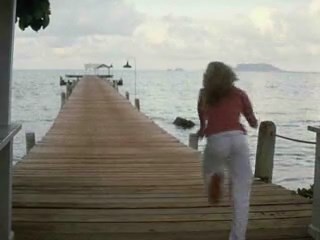}
    \end{subfigure}
    \begin{subfigure}{0.325\linewidth}
        \includegraphics[width=\linewidth]{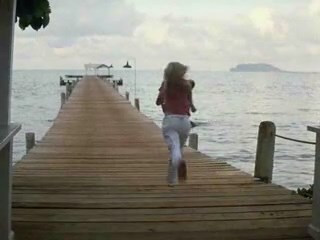}
    \end{subfigure}
    \begin{subfigure}{0.325\linewidth}
        \includegraphics[width=\linewidth]{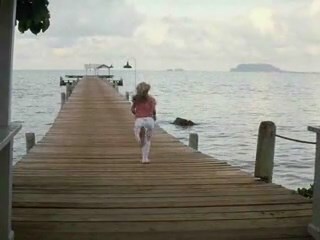}
    \end{subfigure}
    \hfill
    \begin{subfigure}{0.325\linewidth}
        \includegraphics[width=\linewidth]{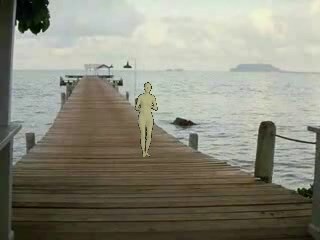}
    \end{subfigure}
    \begin{subfigure}{0.325\linewidth}
        \includegraphics[width=\linewidth]{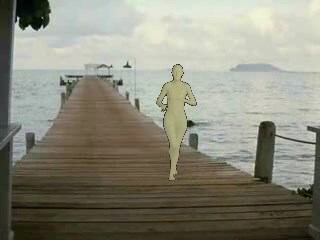}
    \end{subfigure}
    \begin{subfigure}{0.325\linewidth}
        \includegraphics[width=\linewidth]{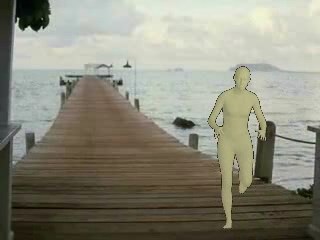}
    \end{subfigure}
    \caption{Example of a video of a woman running towards a pier on the top three images. In the bottom we have an avatar rotated 180 degrees and running from the pier, as we have generated the synthesis backwards.}
    \label{fig:example_3}
\end{figure}

We are also able to generate hybrid content that is semantically different from the original video. For instance, in Fig.~\ref{fig:example_3} we show that we can transform a video of a person running towards a pier in a video where the avatar is actually running from the pier. We are able to do this by introducing a rotation and specifying to the Scene Synthesis module to produce the results backwards. 

%%%%%%%%%%%%%%%%%%%%%%%%%%%%%%%%%%%%%%%%%%%%%%%%%%%%%%%%%%%%%%%%%%%%%%%%%%%%%%%%%%%%%%%%%%%%%%%%%%%%%%%%%%%%%%%%%%%%%%%%%%%%%%%%%%%%%%%%%%%%%%%
\section{Conclusion and Future Work}\label{sec:conclusions}
In this work we explore the vast field that is visual scene understanding and introduce a different paradigm that is visual scene understanding for 3D synthesis, where visual and semantic data extracted from real visual scenes can be leveraged to provide concise descriptions of their underlying scenes and used to recreate 3D virtual environments. We further explore this idea and present a comprehensive architecture for the task of visual scene understanding for 3D synthesis. We address the complexities of analyzing, describing, and recreating scenes in 3D through the proposal of a modular architecture that allows a seamless integration of scene analysis, description, and synthesis. 

For each module of our proposal we explore existing technologies and algorithms that can be employed and show how each module can interact with the other to ensure that the information is always well understood inside all of the modules. We also highlight the benefits of this approach, and explore real-world applications where our proposal can have impact. Additionally, we provide a proof of concept example where the architecture's ability to detect and track humans, estimate and describe their poses, and generate 3D avatars is demonstrated. We combine state-of-the-art computer vision techniques and algorithms to show that it is possible to use our proposal to generate and manipulate visual content with 3D virtual humans performing actions.%, which can be valuable to generate augmented datasets or   

While our work provides a robust foundation for the development of a system or framework, it leaves multiple opportunities for future research and development of associated topics, namely: research on fine-grained scene description supported by the scene graphs that allow precise details about object attributes, environment, scene illumination or group behaviors, which can lead to even more precise descriptions of the scene and ultimately a better and more realistic scene synthesis; Dynamic scene synthesis, where the generated content can incorporate lighting changes; real-time processing for integration in applications such as live broadcasts or mixed reality scenarios. Naturally, other challenges and research opportunities could also be encountered when applying our proposal into more specific application scenarios, ultimately showing that this is a new and challenging task that can see a broad applicability.

\section*{Acknowledgments}
The work was funded by the European Union’s Horizon Europe research and innovation programme under Grant Agreement No 101094831 (Project Converge - Telecommunications and Computer Vision Convergence Tools for Research Infrastructures). Américo was funded by National Funds through the Portuguese funding agency, FCT - Fundação para a Ciência e a Tecnologia, within the PhD grant SFRH/BD/146400/2019


\begin{thebibliography}{4}

\bibitem{cui2023fusing}Cui, X., Khan, D., He, Z. \& Cheng, Z. Fusing surveillance videos and three-dimensional scene: A mixed reality system. {\em Computer Animation And Virtual Worlds}. \textbf{34}, e2129 (2023)
\bibitem{pereira2023visualscene}Pereira, A., Carvalho, P., Pereira, N., Viana, P. \& Côrte-Real, L. From a Visual Scene to a Virtual Representation: A Cross-Domain Review. {\em IEEE Access}. \textbf{11} pp. 57916-57933 (2023)
\bibitem{antoniou2009web}Antoniou, G. \& Harmelen, F. Web ontology language: Owl. {\em Handbook On Ontologies}. pp. 91-110 (2009)
\bibitem{McBride2004}McBride, B. The resource description framework (RDF) and its vocabulary description language RDFS. {\em Handbook On Ontologies}. pp. 51-65 (2004)
\bibitem{sporny2020json}Sporny, M., Longley, D., Kellogg, G., Lanthaler, M. \& Lindström, N. JSON-LD 1.1. {\em W3C Recommendation, Jul}. (2020)
\bibitem{krasin2017openimages}Krasin, I., Duerig, T., Alldrin, N., Ferrari, V., Abu-El-Haija, S., Kuznetsova, A., Rom, H., Uijlings, J., Popov, S., Veit, A. \& Others Openimages: A public dataset for large-scale multi-label and multi-class image classification. {\em Dataset Available From Https://github. Com/openimages}. \textbf{2}, 18 (2017)
\bibitem{deng2009imagenet}Deng, J., Dong, W., Socher, R., Li, L., Li, K. \& Fei-Fei, L. Imagenet: A large-scale hierarchical image database. {\em 2009 IEEE Conference On Computer Vision And Pattern Recognition}. pp. 248-255 (2009)
\bibitem{lin2014microsoft}Lin, T., Maire, M., Belongie, S., Hays, J., Perona, P., Ramanan, D., Dollár, P. \& Zitnick, C. Microsoft coco: Common objects in context. {\em European Conference On Computer Vision}. pp. 740-755 (2014)
\bibitem{jocher2023yolo}Jocher, G., Chaurasia, A. \& Qiu, J. YOLO by Ultralytics.  (2023,1), https://github.com/ultralytics/ultralytics, GitHub repository
\bibitem{kocabas2020vibe}Kocabas, M., Athanasiou, N. \& Black, M. VIBE: Video Inference for Human Body Pose and Shape Estimation. {\em European Conference On Computer Vision (ECCV)}. (2020)
\bibitem{loper2015smpl}Loper, M., Mahmood, N., Romero, J., Pons-Moll, G. \& Black, M. SMPL: A Skinned Multi-Person Linear Model. {\em ACM Transactions On Graphics (TOG)}. \textbf{34}, 248:1-248:16 (2015)
\bibitem{liCvpr22vInpainting}Li, Z., Lu, C., Qin, J., Guo, C. \& Cheng, M. Towards An End-to-End Framework for Flow-Guided Video Inpainting. {\em IEEE Conference On Computer Vision And Pattern Recognition (CVPR)}. (2022)
\bibitem{kuehne2011hmdb51}Kuehne, H., Jhuang, H., Stiefelhagen, R. \& Serre, T. HMDB: A large video database for human motion recognition. {\em 2011 International Conference On Computer Vision (ICCV)}. pp. 2556-2563 (2011)
\bibitem{sowa1992semantic}Sowa, J. \& Others Semantic networks. {\em Encyclopedia Of Artificial Intelligence}. \textbf{2} pp. 1493-1511 (1992)
\bibitem{brock2021high}Brock, A., De, S., Smith, S. \& Simonyan, K. High-performance large-scale image recognition without normalization. {\em International Conference On Machine Learning}. pp. 1059-1071 (2021)
\bibitem{ding2021repmlp}Ding, X., Xia, C., Zhang, X., Chu, X., Han, J. \& Ding, G. RepMLP: Re-parameterizing convolutions into fully-connected layers for image recognition. {\em ArXiv Preprint ArXiv:2105.01883}. (2021)
\bibitem{wang2021swiftnet}Wang, H., Jiang, X., Ren, H., Hu, Y. \& Bai, S. Swiftnet: Real-time video object segmentation. {\em Proceedings Of The IEEE/CVF Conference On Computer Vision And Pattern Recognition}. pp. 1296-1305 (2021)
\bibitem{hu2021learning}Hu, L., Zhang, P., Zhang, B., Pan, P., Xu, Y. \& Jin, R. Learning position and target consistency for memory-based video object segmentation. {\em Proceedings Of The IEEE/CVF Conference On Computer Vision And Pattern Recognition}. pp. 4144-4154 (2021)
\bibitem{wang2021salient}Wang, W., Lai, Q., Fu, H., Shen, J., Ling, H. \& Yang, R. Salient object detection in the deep learning era: An in-depth survey. {\em IEEE Transactions On Pattern Analysis And Machine Intelligence}. (2021)
\bibitem{qin2022idyolo}Qin, L., Shi, Y., He, Y., Zhang, J., Zhang, X., Li, Y., Deng, T. \& Yan, H. ID-YOLO: Real-Time Salient Object Detection Based on the Driver's Fixation Region. {\em IEEE Transactions On Intelligent Transportation Systems}. (2022)
\bibitem{redmon2018yolov3}Redmon, J. \& Farhadi, A. Yolov3: An incremental improvement. {\em ArXiv Preprint ArXiv:1804.02767}. (2018)
\bibitem{hurtik2022polyyolo}Hurtik, P., Molek, V., Hula, J., Vajgl, M., Vlasanek, P. \& Nejezchleba, T. Poly-YOLO: Higher Speed, More Precise Detection and Instance Segmentation for YOLOv3. {\em Neural Computing And Applications}. pp. 1-16 (2022)
\bibitem{johnson2015image}Johnson, J., Krishna, R., Stark, M., Li, L., Shamma, D., Bernstein, M. \& Fei-Fei, L. Image retrieval using scene graphs. {\em Proceedings Of The IEEE Conference On Computer Vision And Pattern Recognition}. pp. 3668-3678 (2015)
\bibitem{gao2018image}Gao, L., Wang, B. \& Wang, W. Image captioning with scene-graph based semantic concepts. {\em Proceedings Of The 2018 10th International Conference On Machine Learning And Computing}. pp. 225-229 (2018)
\bibitem{yang2019autoencoding}Yang, X., Tang, K., Zhang, H. \& Cai, J. Auto-encoding scene graphs for image captioning. {\em Proceedings Of The IEEE Conference On Computer Vision And Pattern Recognition}. pp. 10,685-10,694 (2019)
\bibitem{kim2019dense}Kim, D., Choi, J., Oh, T. \& Kweon, I. Dense relational captioning: Triple-stream networks for relationship-based captioning. {\em Proceedings Of The IEEE Conference On Computer Vision And Pattern Recognition}. pp. 6271-6280 (2019)
\bibitem{koner2021graphhopper}Koner, R., Li, H., Hildebrandt, M., Das, D., Tresp, V. \& Günnemann, S. Graphhopper: Multi-hop Scene Graph Reasoning for Visual Question Answering. {\em International Semantic Web Conference}. pp. 111-127 (2021)
\bibitem{nuthalapati2021lightweight}Nuthalapati, S., Chandradevan, R., Giunchiglia, E., Li, B., Kayser, M., Lukasiewicz, T. \& Yang, C. Lightweight Visual Question Answering using Scene Graphs. {\em Proceedings Of The 30th ACM International Conference On Information \& Knowledge Management}. pp. 3353-3357 (2021)
\bibitem{cherian2022spatio}Cherian, A., Hori, C., Marks, T. \& Le Roux, J. (2.5+ 1) D Spatio-Temporal Scene Graphs for Video Question Answering. {\em ArXiv Preprint ArXiv:2202.09277}. (2022)
\bibitem{nguyen2021deep}Nguyen, M., Nguyen, B. \& Gurrin, C. A Deep Local and Global Scene-Graph Matching for Image-Text Retrieval. {\em ArXiv Preprint ArXiv:2106.02400}. (2021)
\bibitem{schroeder2020structured}Schroeder, B. \& Tripathi, S. Structured query-based image retrieval using scene graphs. {\em Proceedings Of The IEEE/CVF Conference On Computer Vision And Pattern Recognition Workshops}. pp. 178-179 (2020)
\bibitem{mody2022analysis}Mody, S. \& Thakkar, J. Analysis of Image generation using Scene graphs. {\em 5th Joint International Conference On Data Science \& Management Of Data (9th ACM IKDD CODS And 27th COMAD)}. pp. 296-297 (2022)
\bibitem{xue2022deep}Xue, Y., Guo, Y., Zhang, H., Xu, T., Zhang, S. \& Huang, X. Deep image synthesis from intuitive user input: A review and perspectives. {\em Computational Visual Media}. \textbf{8}, 3-31 (2022)
\bibitem{tsai2019video}Tsai, Y., Divvala, S., Morency, L., Salakhutdinov, R. \& Farhadi, A. Video relationship reasoning using gated spatio-temporal energy graph. {\em Proceedings Of The IEEE Conference On Computer Vision And Pattern Recognition}. pp. 10,424-10,433 (2019)
\bibitem{malawade2022spatio}Malawade, A., Yu, S., Hsu, B., Muthirayan, D., Khargonekar, P. \& Al Faruque, M. Spatio-Temporal Scene-Graph Embedding for Autonomous Vehicle Collision Prediction. {\em IEEE Internet Of Things Journal}. (2022)
\bibitem{bilal2022transfer}Bilal, M., Maqsood, M., Yasmin, S., Hasan, N. \& Rho, S. A transfer learning-based efficient spatiotemporal human action recognition framework for long and overlapping action classes. {\em The Journal Of Supercomputing}. \textbf{78}, 2873-2908 (2022)
\bibitem{brehar2021pedestrian}Brehar, R., Muresan, M., Mariţa, T., Vancea, C., Negru, M. \& Nedevschi, S. Pedestrian Street-Cross Action Recognition in Monocular Far Infrared Sequences. {\em IEEE Access}. \textbf{9} pp. 74302-74324 (2021)
\bibitem{patil2022survey}Patil, S. \& Prabhushetty, K. A Survey on Human Action Recognition and Detection Techniques. {\em ICT Analysis And Applications}. pp. 157-165 (2022)
\bibitem{song2021human}Song, L., Yu, G., Yuan, J. \& Liu, Z. Human pose estimation and its application to action recognition: A survey. {\em Journal Of Visual Communication And Image Representation}. \textbf{76} pp. 103055 (2021)
\bibitem{ren2020survey}Ren, B., Liu, M., Ding, R. \& Liu, H. A survey on 3d skeleton-based action recognition using learning method. {\em ArXiv Preprint ArXiv:2002.05907}. (2020)
\bibitem{ferreira2022gan}Ferreira, A., Li, J., Pomykala, K., Kleesiek, J., Alves, V. \& Egger, J. GAN-based generation of realistic 3D data: A systematic review and taxonomy. {\em ArXiv Preprint ArXiv:2207.01390}. (2022)
\bibitem{fu2021single}Fu, K., Peng, J., He, Q. \& Zhang, H. Single image 3D object reconstruction based on deep learning: A review. {\em Multimedia Tools And Applications}. \textbf{80} pp. 463-498 (2021)
\bibitem{gao2022nerf}Gao, K., Gao, Y., He, H., Lu, D., Xu, L. \& Li, J. Nerf: Neural radiance field in 3d vision, a comprehensive review. {\em ArXiv Preprint ArXiv:2210.00379}. (2022)
\bibitem{rabby2023beyondpixels}Rabby, A. \& Zhang, C. BeyondPixels: A Comprehensive Review of the Evolution of Neural Radiance Fields. {\em ArXiv Preprint ArXiv:2306.03000}. (2023)
\bibitem{shi2022deep}Shi, Z., Peng, S., Xu, Y., Liao, Y. \& Shen, Y. Deep generative models on 3d representations: A survey. {\em ArXiv Preprint ArXiv:2210.15663}. (2022)
\bibitem{wang2022human}Wang, K., Zhang, G. \& Yang, J. 3D human pose and shape estimation with dense correspondence from a single depth image. {\em The Visual Computer}. pp. 1-13 (2022)
\bibitem{xu2022surface}Xu, T., Fujita, Y. \& Matsumoto, E. Surface-Aligned Neural Radiance Fields for Controllable 3D Human Synthesis. {\em ArXiv Preprint ArXiv:2201.01683}. (2022)
\bibitem{vidaurre2020fully}Vidaurre, R., Santesteban, I., Garces, E. \& Casas, D. Fully Convolutional Graph Neural Networks for Parametric Virtual Try‐On. {\em Computer Graphics Forum}. \textbf{39}, 145-156 (2020)
\bibitem{yu2022multi}Yu, H., Cheang, C., Fu, Y. \& Xue, X. Multi-view Shape Generation for 3D Human-like Body. {\em ACM Transactions On Multimedia Computing, Communications, And Applications (TOMM)}. (2022)
\bibitem{unity3d}Technologies, U. Unity3D. (https://unity.com/,2023), Accessed on 27 September 2023
\bibitem{unrealengine4}Epic Games, I. Unreal Engine 4. (https://www.unrealengine.com,2023), Accessed on 27 September 2023
\bibitem{blender}Foundation, B. Blender. (https://www.blender.org/,2023), Accessed on 27 September 2023
\bibitem{baillet2018forging}Baillet, A., Murphy, E., Dunn, O. \& Gao, M. Forging a new animation pipeline with USD. {\em ACM SIGGRAPH 2018 Talks}. pp. 1-2 (2018)
\bibitem{zhou2023mtanet}Zhou, W., Dong, S., Lei, J. \& Yu, L. MTANet: Multitask-Aware Network With Hierarchical Multimodal Fusion for RGB-T Urban Scene Understanding. {\em IEEE Transactions On Intelligent Vehicles}. \textbf{8}, 48-58 (2023)
\bibitem{hu2020probabilistic}Hu, A., Cotter, F., Mohan, N., Gurau, C. \& Kendall, A. Probabilistic future prediction for video scene understanding. {\em Computer Vision–ECCV 2020: 16th European Conference, Glasgow, UK, August 23–28, 2020, Proceedings, Part XVI 16}. pp. 767-785 (2020)
\bibitem{hu2021bidirectional}Hu, W., Zhao, H., Jiang, L., Jia, J. \& Wong, T. Bidirectional projection network for cross dimension scene understanding. {\em Proceedings Of The IEEE/CVF Conference On Computer Vision And Pattern Recognition}. pp. 14373-14382 (2021)
\bibitem{Cai_2021_CVPR}Cai, Y., Chen, X., Zhang, C., Lin, K., Wang, X. \& Li, H. Semantic Scene Completion via Integrating Instances and Scene In-the-Loop. {\em Proceedings Of The IEEE/CVF Conference On Computer Vision And Pattern Recognition (CVPR)}. pp. 324-333 (2021,6)
\bibitem{Cao_2022_CVPR}Cao, A. \& Charette, R. MonoScene: Monocular 3D Semantic Scene Completion. {\em Proceedings Of The IEEE/CVF Conference On Computer Vision And Pattern Recognition (CVPR)}. pp. 3991-4001 (2022,6)
\bibitem{Nie_2021_CVPR}Nie, Y., Hou, J., Han, X. \& Niessner, M. RfD-Net: Point Scene Understanding by Semantic Instance Reconstruction. {\em Proceedings Of The IEEE/CVF Conference On Computer Vision And Pattern Recognition (CVPR)}. pp. 4608-4618 (2021,6)
\bibitem{behley2021towards}Behley, J., Garbade, M., Milioto, A., Quenzel, J., Behnke, S., Gall, J. \& Stachniss, C. Towards 3D LiDAR-based semantic scene understanding of 3D point cloud sequences: The SemanticKITTI Dataset. {\em The International Journal Of Robotics Research}. \textbf{40}, 959-967 (2021)
\bibitem{johnson2018image}Johnson, J., Gupta, A. \& Fei-Fei, L. Image generation from scene graphs. {\em Proceedings Of The IEEE Conference On Computer Vision And Pattern Recognition}. pp. 1219-1228 (2018)
\bibitem{olszewska2011ontology}Olszewska, J. \& McCluskey, T. Ontology-coupled active contours for dynamic video scene understanding. {\em 2011 15th IEEE International Conference On Intelligent Engineering Systems}. pp. 369-374 (2011)
\bibitem{kenfack2020robotvqa}Kenfack, F., Siddiky, F., Balint-Benczedi, F. \& Beetz, M. Robotvqa—a scene-graph-and deep-learning-based visual question answering system for robot manipulation. {\em 2020 IEEE/RSJ International Conference On Intelligent Robots And Systems (IROS)}. pp. 9667-9674 (2020)
\bibitem{amodeo2022og}Amodeo, F., Caballero, F., Díaz-Rodríguez, N. \& Merino, L. OG-SGG: ontology-guided scene graph generation—a case study in transfer learning for telepresence robotics. {\em IEEE Access}. \textbf{10} pp. 132564-132583 (2022)
\bibitem{cong2021spatial}Cong, Y., Liao, W., Ackermann, H., Rosenhahn, B. \& Yang, M. Spatial-temporal transformer for dynamic scene graph generation. {\em Proceedings Of The IEEE/CVF International Conference On Computer Vision}. pp. 16372-16382 (2021)
\bibitem{xu2022meta}Xu, L., Qu, H., Kuen, J., Gu, J. \& Liu, J. Meta spatio-temporal debiasing for video scene graph generation. {\em European Conference On Computer Vision}. pp. 374-390 (2022)
\bibitem{li2022dynamic}Li, Y., Yang, X. \& Xu, C. Dynamic scene graph generation via anticipatory pre-training. {\em Proceedings Of The IEEE/CVF Conference On Computer Vision And Pattern Recognition}. pp. 13874-13883 (2022)
\bibitem{Dobbie2009}Dobbie, G. \& Ling, T. Object Relationship Attribute Data Model for Semi-structured Data. {\em Encyclopedia Of Database Systems}. pp. 1940-1941 (2009)


\end{thebibliography}
\end{document}